# Asymmetric Lesion Detection with Geometric Patterns and CNN-SVM Classification

M. A. Rasel[1], Sameem Abdul Kareem[1], Zhenli Kwan[2], Nik Aimee Azizah Faheem[2], Winn Hui Han[2], Rebecca Kai Jan Choong[2], Shin Shen Yong[2], Unaizah Obaidellah[1,*]

[1]Department of Artificial Intelligence, Faculty of Computer Science and Information Technology, Universiti Malaya, Kuala Lumpur, 50603, Malaysia
[2]Division of Dermatology, Department of Medicine, Faculty of Medicine, Universiti Malaya, Kuala Lumpur, 50603, Malaysia
[*]Corresponding Author: Unaizah Obaidellah. Email: unaizah@um.edu.my

**Abstract**

In dermoscopic images, which allow visualization of surface skin structures not visible to the naked eye, lesion shape offers vital insights into skin diseases. In clinically practiced methods, asymmetric lesion shape is one of the criteria for diagnosing Melanoma. Initially, we labeled data for a non-annotated dataset with symmetrical information based on clinical assessments. Subsequently, we propose a supporting technique—a supervised learning image processing algorithm—to analyze the geometrical pattern of lesion shape, aiding non-experts in understanding the criteria of an asymmetric lesion. We then utilize a pre-trained convolutional neural network (CNN) to extract shape, color, and texture features from dermoscopic images for training a multiclass support vector machine (SVM) classifier, outperforming state-of-the-art methods from the literature. In the geometry-based experiment, we achieved a 99.00% detection rate for dermatological asymmetric lesions. In the CNN-based experiment, the best performance is found 94% Kappa Score, 95% Macro F1-score, and 97% weighted F1-score for classifying lesion shapes (Asymmetric, Half-Symmetric, and Symmetric).

## 1. Introduction

Dermatological asymmetry, a cornerstone in skin lesion assessment, refers to disparities observed in the shape, size, or color of moles or lesions [1, 2, 3]. In dermatology, careful examination of the lesion shape is critical, especially when it comes to the possibility that lesions are cancerous, such as Melanoma. Asymmetry, an important parameter delineated in the ABCDE rule [1], signifies asymmetry between halves of a mole or lesion regarding shape, size, or color. The dermatological three-point-checklist for early skin cancer detection has showcased remarkable sensitivity in identifying Melanoma [2]. The presence of "asymmetry of color and structure in one or two perpendicular axes", stands as the initial criterion of this checklist [2]. Moreover, the CASH method for Melanoma detection integrates the asymmetry criterion [3]. In this method, asymmetry evaluation entails scrutinizing lesions within a plane bisected by two axes set at 90°, assigning a score ranging from 0 to 2 based on the number of axes exhibiting asymmetry in shape, color, or structure. While asymmetrical lesions may occasionally prove benign, they often necessitate further assessment due to their association with potential malignancy. This discrepancy serves as an indicative feature of potential malignancy, as illustrated in **Fig. 1**, which showcases both asymmetric and symmetric skin lesions from the ISIC2016 dataset [4].

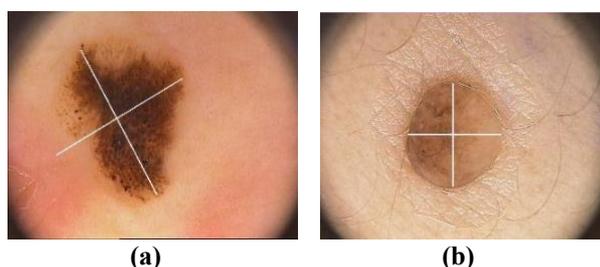

**(a)**      **(b)**
**Fig. 1. (a)** Asymmetric (do not match); and **(b)** Symmetric (almost match) lesions are from the ISIC2016 dataset [4].

The requirement for precise and prompt evaluation has driven the incorporation of artificial intelligence in medical settings, resulting in inventive approaches that enable the computerized examination of dermoscopic images. In recent work, researchers have used CNNs to create advanced algorithms capable of distinguishing between benign and malignant lesions. These CNN-based algorithms have revolutionized early skin cancer detection by quickly identifying potential malignancies upon training on large dermatological image datasets. Although a considerable amount of research has been conducted to automatically diagnose skin cancer using dermoscopic images, only a few studies have focused on dermoscopic features such as lesion asymmetry. Utilizing CNNs to identify asymmetric skin lesions can bridge this research gap. On the other hand, an extensive database of training data is essential to acquire CNN effectiveness. Since there are only a handful of datasets in the literature dedicated to training models with asymmetric lesions, more labeled data are required to enhance effectiveness. This data requirement also provides an opportunity to contribute to lesion asymmetry research by introducing a new annotated dataset. Sometimes, data labeling based on clinical experts' opinions is also a challenging task. During data labeling, differences in expert judgments resulting from the intricate internal architecture of lesion regions motivate the exploration of computer vision-based methods to assist in lesion shape identification. Furthermore, leveraging a CNN for feature extraction from medical images, researchers have successfully captured intricate patterns and textures indicative of various diseases. Subsequent training of a classifier on these extracted features enables accurate disease or symptom categorization. This novel method ensures uniformity between assessments and improves the speed and precision of skin lesion analysis, allowing for earlier treatments. This technological advancement presents another opportunity to implement an automated method for detecting asymmetric lesions.

Based on the aforementioned research gaps, a non-annotated dataset becomes an annotated dataset in this research. Then, the methodology of lesion shape analysis in this study is divided into two parts. The first part proposes a pixel count-based imaging technique aimed at facilitating the understanding of lesion shape. The primary objective of this component is to aid experts in circumventing opinion variations and providing labeled data for CNN training when a sufficient volume of labeled data is required. In the second part, a CNN is proposed to autonomously detect asymmetric lesions to improve the understanding of skin conditions.

This combination of state-of-the-art technology and dermatological knowledge could greatly reduce the burden on dermatologists in identifying skin-related diseases to improve patient outcomes. CNN-based automated analysis is essential for reliable, accurate, and consistent evaluation, potentially outperforming manual methods and allowing for timely interventions in medicine that may save lives.

The remainder of this paper is organized as follows: Section 2 reviews related work in the field of dermoscopic image analysis. Section 3 describes the methodology used for asymmetric analysis. Section 4 presents the experimental results and discusses the findings and their implications. Finally, Section 5 concludes the paper and suggests directions for future research.

## 2. Related Works

The analysis of dermatological asymmetry has been the subject of investigation in various research works found in the literature. **Table 1** provides a systematic review of such research, revealing that only a limited number of studies have focused on this feature extraction process. This could be attributed to the perception that the individual lesion shape alone holds little diagnostic value for disease diagnosis. However, it is crucial to recognize that lesion shape plays a leading role in diagnosing Melanoma, as it serves as a vital indicator for accurate identification.

**Table 1**
A review of dermatological asymmetric lesion analysis research works.

| Authors | Aim: Asymmetry | Database | Methodology | Evaluation | (c9) Limitation |
|---|---|---|---|---|---|
| Lorentzen et al., 2001[5] | Classification | Own | Latent class analysis | Sensitivity = 77-92% | Supervised, Single dataset |
| Ng et al., 2005 [6] | Analysis, Classification | Own | Image Processing, Backpropagation | Accuracy = 80% | Limited data, Supervised |
| Sirakov et al., 2011 [7] | Analysis, Classification | EDRA | Image Processing | Accuracy = 95.00% | Supervised, Limited data |
| Chakravorty et al., 2016[8] | Classification | PH2 | Image Processing, Machine Learning | Accuracy = 87.00% | Single dataset, Semiautomatic |
| Milczarski, 2017[9] | Analysis, Classification | PH2 | Image Processing | Accuracy = 95.80% | Single dataset, Supervised |
| Sancen-Plaza et al., 2018 [10] | Analysis, Classification | Lee, PH2 | Normalized E-Factor | Sensitivity = 59.62, Specificity = 85.8% | Supervised, Limited data |
| Ali et al., 2020[11] | Analysis, Classification | ISIC2018 | Image Processing, Machine Learning | Accuracy = 80.00% | Single dataset, Semiautomatic |
| Damian et al., 2021[12] | Classification | Med-Node, PH2 | Image Processing, Artificial Neural Network | Not mentioned | Limited data, Complex |
| Zhang and Guo, 2021 [13] | Analysis, Classification | Not mentioned | Image Processing, Pattern Recognition | Not mentioned | Supervised |
| Talavera-Martínez et al., 2022[14] | Classification | SymDerm | Pre-trained CNN, Transfer Learning | Accuracy = 64.50% | Low Accuracy |

Lorentzen et al. [5] assessed the asymmetry in pigmented skin lesions which held significant importance in the diagnosis of malignant Melanoma, as outlined in various dermatological diagnostic rules. However, its subjective nature can lead to variability among observers, highlighting the need for objective evaluation methods. This study aimed to enhance sensitivity in detecting axis (a-) symmetry through latent class analysis (LCA), analyzing ratings from expert dermatologists on 232 pigmented lesions. The findings revealed varying sensitivities for different levels of asymmetry, with Melanomas exhibiting higher asymmetry compared to other lesions. LCA proved effective in minimizing observer dependence and providing more accurate lesion classification, emphasizing the crucial role of asymmetry assessment in Melanoma diagnosis and risk stratification.

Ng et al. [6] introduced an adaptive fuzzy method that employs symmetric distance (SD) to assess lesions with fuzzy borders. By incorporating various SD variations and utilizing a backpropagation neural network, the approach achieved improved discriminative capability. Application of the method to digitized images obtained from the Lesion Clinic inVancouver, Canada, showcases its ability to accurately classify asymmetric lesions, using a set of 120 digital images, comprising 60 symmetric and 60 asymmetric lesions.

Sirakov et al. [7] introduced a method and tool for the automated extraction of skin lesion boundaries, crucial for symmetry and area calculations. An image enhancement technique was employed to prepare each image for active contour (AC) evolution. The AC method then automatically delineated the lesion boundary using a minimal boundary box, facilitating symmetry measurement. Additionally, the area of the lesion was computed. Consequently, the lesions were plotted as points in a 2D space representing area symmetry, aiding in the identification of cancerous lesions. To validate these theoretical concepts, experiments were conducted using 51 skin lesion images, and statistical analysis was employed to assess the accuracy of boundary extraction compared to ground truth.

A study by Chakravorty et al. [8] focused on Melanoma asymmetry as a characteristic of early diagnosis. The authors employed the Kullback-Leibler Divergence of the color histogram and Structural Similarity metric to measure irregularities in color and structure distribution within the lesion area. Several classifiers utilizing these features were evaluated on a dataset, demonstrating improved asymmetry classification compared to existing literature. However, it is important to note that the method used in this experiment was not fully automated, and for evaluation purposes, only one dataset was used, highlighting a limitation.

In a paper by Milczarski [9], a dermatological asymmetry measure in hue (DASMHue) was introduced and discussed as part of the three-point checklist for skin lesions. The focus of this paper was on the hue distribution asymmetry of segmented skin lesions, and new dermatological asymmetry measures based on hue distribution were defined. The presented DASMHue measure showed stronger overestimating results but achieved a better total ratio (95.8%) of correctly and overestimated cases compared to considering shape alone, based on results from the dataset. However, it's worth noting that this experiment didn't involve many datasets or any automation process.

Sancen-Plaza et al. [10] focused on developing an algorithm to quantify asymmetry in skin lesions, a key component in computer aided diagnosis Systems for early Melanoma detection. The algorithm divided lesion images into segments and calculated discrete compactness values using Normalized E-Factor (NEF). By measuring the sum of squared differences between NEF values and their corresponding opposites, asymmetry values were obtained. The algorithm's efficacy was evaluated using two public skin cancer databases. In first dataset, strong correlations were found between dermatologists' diagnoses and asymmetry values, especially for likely Melanomas. In the second dataset, the algorithm demonstrated promising sensitivity and specificity compared to dermatologist assessments. This approach, based on simple image digital features, provided a stable and accurate measure of asymmetry in skin lesions, crucial for early Melanoma detection.

A paper by Ali et al. [11] addressed the need for an objective computer vision system to aid in the early detection of Melanoma by evaluating asymmetry, color variegation, and diameter. The proposed approach involved training a decision tree on extracted asymmetry measures to predict the asymmetry of new skin lesion images. Suspicious colors for color variegation were derived, and Feret's diameter was used to determine the lesion's diameter. The decision tree achieved an 80% accuracy in determining asymmetry, while the number of suspicious colors and diameter values was objectively identified. However, this percentage rate of accuracy couldn't demonstrate any promising achievement.

A paper by Damian et al. [12] introduced a feedforward neural network (FFN) with the Levenberg-Marquardt Backpropagation (LMBP) training algorithm to analyze skin lesions. The model utilized different combinations of inputs and desired outputs related to skin lesion types, databases, and asymmetry computation methods. The FFN-LMBP model was validated and tested on 24 images each, focusing on the asymmetry feature extracted using geometric characteristics and histogram projection algorithms. Two datasets were used for skin lesion detection in this study. However, the proposed idea was presented briefly, leaving the implementation somewhat challenging.

Zhang and Guo [13] introduced an approach that integrates the ABCD rule with clinical characteristics of Melanoma in dermoscopic images to differentiate between benign and malignant melanocytes. Initially, the method employs image processing and pattern recognition techniques to obtain and preprocess dermoscopic images, extracting feature points from the images. Subsequently, both the microscopic and macroscopic asymmetry of the images are analyzed. Finally, a symmetry score is determined using TDS (total dermoscopic score) to assess the benign or malignant nature of melanocytes.

A study by Talavera-Martínez et al. [14] proposed a novel approach using deep learning techniques to classify skin lesion symmetry in dermoscopic images. A CNN model was employed to classify lesions as fully asymmetric, symmetric concerning one axis, or symmetric concerning two axes. The introduction of a new dataset with 615 labeled skin lesions and the evaluation of transfer learning and traditional learning-based methods contributed to the development of a simple, robust, and efficient classification pipeline, outperforming traditional approaches and pre-trained networks with a weighted-average F1-score of 64.5%. However, the proposed algorithm could not demonstrate any significant achievement in the analysis of dermoscopic asymmetry while maintaining a relatively complex algorithm.

These studies focusing on shape-based analysis of asymmetry lesions have made significant contributions. However, as of the current state of research, there has been limited exploration of deep learning methods in this field, and no existing work has offered a detailed analytical explanation for detecting dermatological asymmetric lesions. Besides, only the PH2 [15] dataset offers ground truth about skin lesion shape. Additionally, there is a lack of geometry-based lesion shape analysis that would facilitate comprehension for non-experts to establish the ground truth regarding the lesion shape types (either asymmetric or symmetric). Here, these gaps are aimed to be addressed by proposing a pixel-based geometric algorithm that will not only enhance the understanding of dermatological shape types for experts but also provide accessible explanations to non-experts. By incorporating deep learning techniques and detailed analytical explanations, this study aims to make substantial advancements in the detection and comprehension of dermatological asymmetric lesions.

## 3. Materials and Method

### 3.1 Data Acquisition

The acquisition of data for the dermoscopic symmetrical analysis research involved a diligent process aimed at gathering comprehensive information regarding lesion symmetry within two datasets sourced from the PH2 [15] and ISIC2016 [4] databases.

**PH2:** The PH2 dataset, developed by Hispano Pedro Hospital, Portugal, stands as the sole publicly available annotated dermoscopic dataset where skin lesion shapes are carefully classified into three distinct categories: symmetric (117 images), half-symmetric (31 images), and asymmetric (52 images). **(a2)** Each image within this dataset provides ground truth for the segmented area of the lesion in a binary mask. In preparation for this research, each of the 200 images comprising the PH2 dataset underwent vertical and horizontal mirroring once. This augmentation technique was essential as the original 200 images were insufficient to effectively train a CNN. By expanding the dataset through mirroring, we increase the total number of dermoscopic images from 200 to 600. This extended dataset is called the augmented PH2 (APH2) dataset, formed a crucial component of our research endeavors.

**ISIC2016:** The ISIC2016 dataset comprises 1279 dermoscopic images accompanied by segmented masks delineating the lesion area in binary format (black and white images). However, this dataset lacked ground truth about lesion symmetry. To address this gap, we assembled a multidisciplinary team consisting of five clinical experts proficient in dermatology. Leveraging their expertise, we systematically annotated the 1279 images to denote the symmetry characteristics of each lesion. These annotations categorize lesions as symmetric (268 images), half-symmetric (344 images), and asymmetric (667 images), providing essential ground truth for our subsequent analysis. To deal with the variability among the experts' decisions when annotating the dataset ISIC2016, the voting system of [14] was used. Our data collection methodology extends beyond mere visual inspection of the images. In addition to subjective evaluations by our expert team, we conducted detailed analyses of both the original high-resolution images and their corresponding segmented masks, which were provided as part of the ISIC2016 dataset. This comprehensive approach ensures robust labeling of lesion symmetry, enhancing the reliability and accuracy of our dataset. Furthermore, to augment our manual annotations, we developed an imaging algorithm designed to quantitatively assess lesion symmetry based on pixel counts within the segmented masks. This algorithm splits each lesion into four distinct sections, calculating the proportion of pixels within each segment relative to the total lesion area. The detailed methodology of this pixel count-based approach is described in Section 5 of this manuscript. By combining manual annotations from clinical dermatologists with computational techniques, we provide ground truths about lesion shape and introduce a methodology for objectively quantifying symmetry characteristics. This comprehensive dataset and methodology form the foundation of our research, enabling rigorous analysis and exploration of dermoscopic lesion symmetry.

Notably, during the data labeling process for ISIC2016, we observed variations in expert opinions regarding lesion shape attributed to dermoscopic structures. Similarly, a non-dermoscopic dataset called SymDerm [14] encountered a similar issue. However, they addressed this "disagreement issue" by utilizing a voting system involving three experts. In addressing such challenges, the proposed pixel-count-based imaging algorithm may serve as a valuable tool to reinforce expert opinions. Later these two datasets: APH2 and ISIC2016 with the ground truth are used to train (95%) and test (5%) the proposed CNN model **(details in Section 6).**

### 3.2 Evaluation Metrics

The experimental results of the classifiers will report these metrics - Precision, Recall, F1-Score, **(c22)** Kappa Score (K), and Accuracy. The experimental datasets containing more than two classes and imbalance data will also be reported. To calculate the average evaluation metrics for the three classes (Asymmetric, Half-Symmetric, and Symmetric), the Macro F1-Score (M-F1) and Weighted F1-Score (W-F1) will be counted. Mathematically they are presented as-

$$Precision = \frac{TP}{TP + FP} \quad (1)$$

$$Recall = \frac{TP}{(TP + FN)} \quad (2)$$

$$F1 - Score = \frac{2TP}{2TP + FP + FN} \quad (3)$$

$$M - F1 = \frac{\sum(F1 - Score\,of\,each\,class)}{no.\,of\,class} \quad (4)$$

$$W - F1 = \frac{\sum(no.\,of\,each\,class\,sample \times each\,class\,F1)}{no.\,of\,total\,samples} \quad (5)$$

$$Kappa\ Score\ (K) = (P_o - P_e)/(1 - P_e) \quad (6)$$

$$Accuracy = \frac{TP + TN}{P + N} \quad (7)$$

For these evaluation metrics, TP is true positive, TN is true negative, FN is false negative, FP is false positive, P is actual positive, and N is actual negative number of the image data. For $K$, $P_o$ is the probability of agreement and $P_e$ is the probability of random agreement for each class. Higher scores of $K$ reflect stronger agreement and better model performance.

### 3.3 Analyzing of Skin Lesion Shape

To analysis skin lesion shape for supporting the decision of clinical experts, a new pixel-count-based imaging approach is proposed called the Geometry Shape-Based Asymmetry Analysis (GSAA). Input for the GSAA is a binary lesion image. Since both PH2 and ISIC2016 have binary lesion images, generating a binary mask for each image is not required. **Fig. 2** shows two images from PH2 and ISIC2016 and their corresponding binary ones.

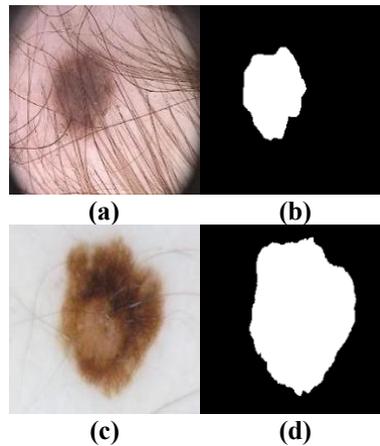

**Fig. 2.** Dermoscopic images (a, c) and corresponding binary images of lesions (b, d), respectively from PH2 (a, b) and ISIC2016 (c, d) datasets.

Each binary (mask) image is split into 4 portions using two perpendicular axes at the lesion centroid. As a result, 4 different split images are found in a single image. **Fig. 3** shows the dividing process of the binary image. Those 4 split images are considered A (segmented area: right side bottom corner), B (segmented area: left side bottom corner), C (segmented area: left side up corner), and D (segmented area: right side up corner).

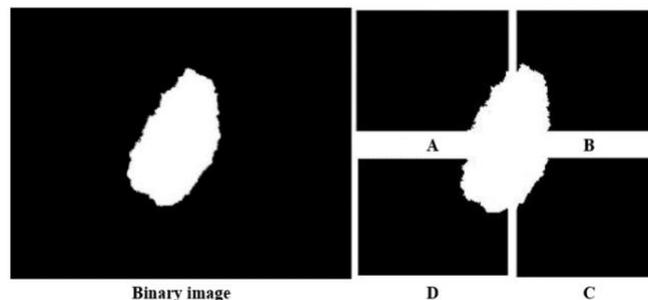

**Fig. 3.** A binary image (from PH2) is divided into 4 split images.

For A, B, C, and D, the number of white pixels is counted separately as they represent the lesion area. Then, their total white pixel numbers are divided by each other to compare the area. Usually, the skin lesions do not have a smooth border and exact symmetrical shape including color and structure [6]. Naturally, the lesion shapes are unsmooth and irregular. Therefore, any two split images are not expected to have the same pixel number of white areas. There might be a fraction of the quotient if we divide the number of the white pixels of 2 split images by each other. Therefore, three conditions are needed to form based on arithmetic assumptions.

$$\text{If, } \frac{any\ split\ image's\ total\ white\ pixel}{another\ split\ image's\ total\ white\ pixel} = 0.90\ to\ 1.10\ (it\ will\ be\ considered\ as\ 1)$$

$$\text{If, } \frac{any\ split\ image's\ total\ white\ pixel}{another\ split\ image's\ total\ white\ pixel} < 0.90\ (it\ will\ be\ considered\ as\ 0)$$

$$\text{If, } \frac{any\ split\ image's\ total\ white\ pixel}{another\ split\ image's\ total\ white\ pixel} > 1.10\ (it\ will\ be\ considered\ as\ 0)$$

Based on these three conditions, lesion binary images are divided from PH2 and ISIC2016 datasets. The division process is followed-

$$Ap/Bp = 0\ or\ 1,\ Ap/Dp = 0\ or\ 1,\ Bp/Cp = 0\ or\ 1,\ and\ Cp/Dp = 0\ or\ 1.$$

Here, Ap, Bp, Cp, and Dp store the total number of white pixels in the split images A, B, C, and D. If a minimum of one quotient and a maximum of two quotients of these four operations become 1, then the lesion will be Symmetric in 1 axis (Half-Symmetric). If any three quotients of these four operations become 1, then the lesion will be Symmetric. If all quotients of these four operations become 0, then the lesion will be Asymmetric. The algorithm of above conditions for lesion shape analysis in binary image is presented below-

> ***Step 1:*** *Take the four (A, B, C, and D) split images from the binary image.*
> ***Step 2:*** *Count the number of white pixels in each part and store them as Ap, Bp, Cp, and Dp.*
> ***Step 3:*** *Calculate the quotients of the white pixel counts for each pair of parts (Ap/Bp, Ap/Dp, Bp/Cp, Cp/Dp).*
> ***Step 4:*** *For each quotient:*
>     *If the quotient is between 0.9 and 1.1, set it to 1; otherwise, set it to 0.*
> ***Step 5:*** *Check the quotients:*
>     *If minimum one quotient and maximum two quotients are 1:*
>         *Return "Half-Symmetric".*
>     *If minimum three quotients are 1:*
>         *Return "Symmetric".*
>     *If all quotients are 0:*
>         *Return "Asymmetric".*

**Table 2** shows the shape analysis results for 5 samples out of 200 samples from PH2. These analytical results match with the ground truth at 99.00% as presented in **Table 3** as a confusion matrix (CM). The CM is a graphical presentation of predicted value vs. ground truth value. The GSAA approach misclassified 2 out of 200 images due to the presence of artifacts and incompleteness (part of the lesion captured).

**Table 2**
GSAA is applied on the PH2 dataset.

| Lesion Image | Ap/Cp | Bp/Dp | Ap/Bp | Cp/Dp | Symmetric | Half-Symmetric | Asymmetric | Ground Truth |
|---|---|---|---|---|---|---|---|---|
| 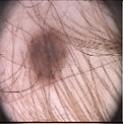 | 1 | 1 | 1 | 1 | Yes | No | No | Symmetric |
| 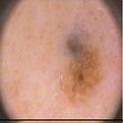 | 0 | 0 | 0 | 0 | No | No | Yes | Asymmetric |
| 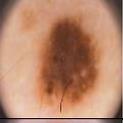 | 0 | 0 | 1 | 1 | No | Yes | No | Half-Symmetric |
| 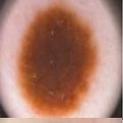 | 1 | 1 | 1 | 1 | Yes | No | No | Symmetric |
| 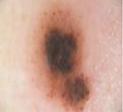 | 0 | 0 | 0 | 0 | No | No | Yes | Asymmetric |

**Table 3**
The CM of the GSAA vs. the ground truth (PH2).

|  |  | Ground Truth | | |
|---|---|---|---|---|
|  |  | **Symmetric** | **Half-Symmetric** | **Asymmetric** |
| **GSAA's Prediction** | **Symmetric** | 116 | 1 | 0 |
|  | **Half-Symmetric** | 1 | 30 | 0 |
|  | **Asymmetric** | 0 | 0 | 52 |

Again, the proposed imaging technique is applied to the ISIC2016 dataset. After applying the GSAA to this dataset, 660 asymmetric, 341 half-symmetric, and 278 symmetric lesion images are found. Based on the ground truth, 12 images are misclassified out of 1279 images due to artifacts and incompleteness, like the results from the PH2 dataset. This is another indication of the GSAA's success, and a significant contribution to the ISIC2016 dataset is the division of skin lesions into three classes. These contributions are clinically approved and geometrically (shape analysis in a mathematical way) analyzed and verified. **Table 4** shows the lesion shape analysis results for random 5 samples. These outputs match the ground truth with an accuracy of 99.06%, as presented in the CM in **Table 5**.

**Table 4**
GSAA is applied on the ISIC2016 dataset.

| Lesion Image | $A_p/C_p$ | $B_p/D_p$ | $A_p/B_p$ | $C_p/D_p$ | Symmetric | Half-Symmetric | Asymmetric | Ground Truth |
|---|---|---|---|---|---|---|---|---|
| 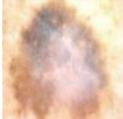 | 0 | 0 | 0 | 0 | No | No | Yes | Asymmetric |
| 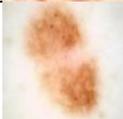 | 1 | 1 | 0 | 0 | No | Yes | No | Half-Symmetric |
| 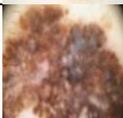 | 0 | 0 | 0 | 0 | No | No | Yes | Asymmetric |
| 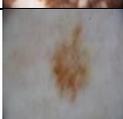 | 0 | 0 | 1 | 1 | No | Yes | No | Half-Symmetric |
| 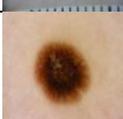 | 1 | 1 | 1 | 1 | Yes | No | No | Symmetric |

**Table 5**
The CM of the GSAA vs. the ground truth (ISIC2016).

|  |  | Ground Truth (five experts provided) | | |
|---|---|---|---|---|
|  |  | Symmetric | Half-Symmetric | Asymmetric |
| **GSAA's Prediction** | Symmetric | 267 | 4 | 7 |
|  | Half-Symmetric | 1 | 340 | 0 |
|  | Asymmetric | 0 | 0 | 660 |

**(a5)** To evaluate GSAA, Eq. (6) and Eq. (7) are applied on Table 3 and Table 5. The GSAA achieved Kappa score 98.2% and accuracy 99.0% for PH2; and Kappa score 98.5% and accuracy 99.1% for ISIC2016. However, for the next part of this research (CNN-based lesion classification), the output of GSAA is not used. Since PH2 and ISIC2016 have clinical-based ground truth, which are more reliable than GSAA generated ground truth. For this reason, the clinical-based ground truth of PH2 and ISIC2016 are used to train and evaluate the proposed CNN. The proposed GSAA technique is beneficial when there is an insufficient amount of annotated data available for training a deep learning model. This method can be utilized to generate ground truth labels for non-annotated datasets, thereby creating an annotated dataset.

### 3.4 Lesion Shape Classification Using CNN

Pre-trained models can save training time and resources, as well as improve performance and accuracy of a classification problem. There are several pre-trained networks that have gained popularity. Most of these have been trained on the ImageNet dataset, which has 1000 object categories and 1.2 million training images [16]. "ResNet18 (71 layers), ResNet50 (177 layers), and ResNet101 (347 layers)" are three such CNN models and can be reused for the classification of skin lesion images. The numbers of learnable parameters for them are respectively 11.6M, 25.5M, and 44.6M [17]. The first layer defines the input dimensions. Usually, each CNN has different input size requirements for the input image. These three used in this experiment require image input that is 224-by-224-by-3. The intermediate layers make up the bulk of CNN. These are a series of convolutional layers, interspersed with rectified linear units (ReLU) and max-pooling layers [18]. Following these layers are 3 fully-connected layers. The final layer is the classification layer, and its properties depend on the classification task. Here, the CNN models that are employed were trained to solve a 1000-way classification problem. Thus, the classification layer has 1000 classes from the ImageNet dataset. However, the classification layer for this shape analysis research has three classes (Asymmetric, Half-Symmetric, and Symmetric) from the APH2 and ISIC2016 datasets. For the training and validation, 95% (75% and 20%) of APH2 and ISIC2016 are randomly selected. For testing, 5% of APH2 (31 images) and ISIC2016 (63 images) are used and not involved in training and validation process.

## 3.5 Extract Training Features Using CNN

Each layer of a CNN produces a response, or activation, to an input image. However, there are only a few layers within a CNN that are suitable for image feature extraction. In several independent experiments, ResNet18, ResNet50, and ResNet101 are each employed separately to extract features from the dermoscopic images. ResNet models offer advantages due to their capacity to mitigate the vanishing gradient issue within the lower layers of the architecture, as well as their potential for extensive scaling [19]. These characteristics enable the model to be adjusted to suit the specific classification task. The first layer of these networks has learned filters for capturing blob and edge features. These "primitive" features are then processed by deeper network layers, which combine the early features to form higher level image features. These higher-level features are better suited for recognition tasks because they combine all the primitive features into a richer image representation [20].

We extract features from one of the deeper layers of the pre-trained model using the activations method. The deeper layers of the network, those closer to the output, capture more complex and abstract features compared to the shallower layers. The "activations method" refers to capturing the output (activations) of a specific layer in the network when an image is passed through it. These activations represent the features learned by that layer. Selecting which of the deep layers to choose is a design choice, but typically starting with the layer right before the classification layer is a good place to start. In the ResNet models, this layer is named 'fc1000' (fully connected layer). Training features are extracted using that layer. **Fig. 4** shows the input layers to fully connected layers of ResNet18, ResNet50 and ResNet101.

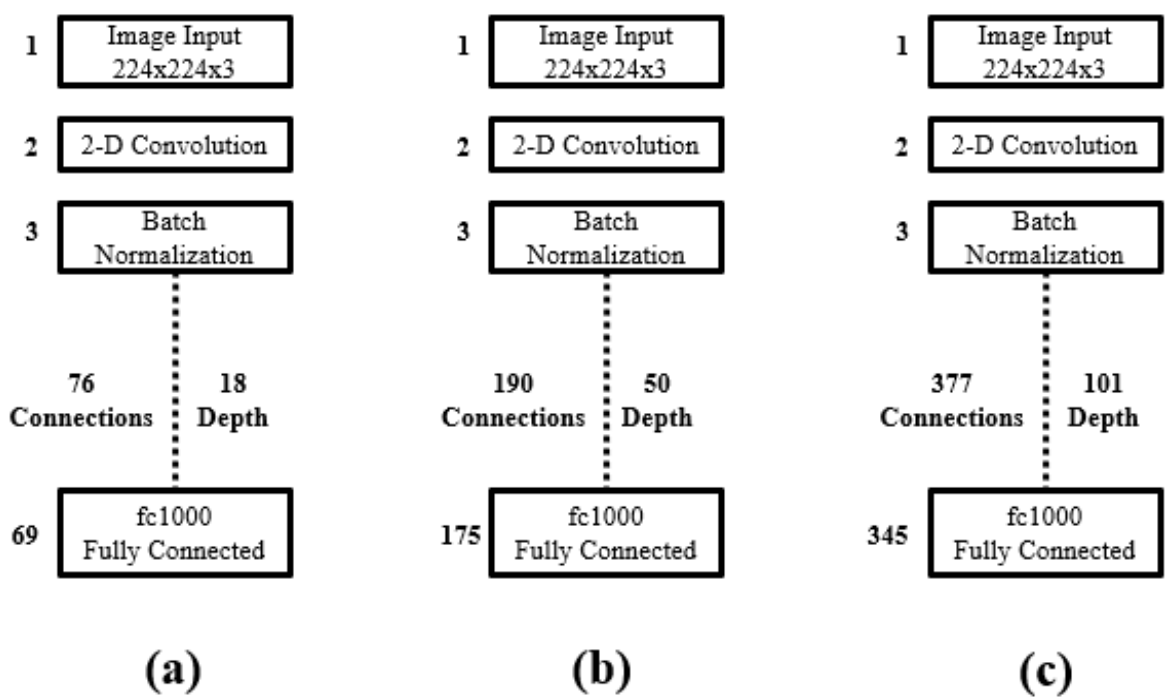

**Fig. 4.** (a), (b), and (c) are briefly presented architecture of ResNet18, ResNet50, and ResNet101.

The 'MiniBatchSize' is set 32 to ensure that the model and image data fit into the system's GPU memory. Also, the activation output is arranged as columns. This helps speed up the multiclass linear SVM training that follows.

### 3.6 Training the Multiclass SVM Classifier Using ResNet Features

The Support Vector Machine (SVM) is a supervised learning algorithm used for classification tasks. In a multiclass setting, SVM extends its binary classification (BC) capability to handle multiple lesion classes. SVM does so by training several binary classifiers on ResNet extracted features, one for each class, and then combining their outputs to make multiclass predictions. Stochastic Gradient Descent (SGD) is an optimization algorithm commonly used to train machine learning models. SGD works by updating the model's parameters iteratively to minimize a loss function [21].

In the context of SVM, SGD is used to optimize the classifier's parameters, such as the weights assigned to different features. The multiclass SVM classifier is trained using Error-Correcting Output Codes (ECOC). By setting the 'Learners' parameter to 'Linear', the process used a linear SVM as the base learner for each binary classifier within the ECOC framework. Here, $K(K – 1)/2$ binary SVM models are applied using the one-versus-one (ovo) coding design, where $K$ is the number of unique class labels.

Ovo is a heuristic method for using BC algorithms for multi-class classification which splits a multi-class classification dataset into BC problems [22]. This approach splits the dataset into one dataset for each class versus every other class. **Fig. 5** shows the multiclass SVMs classification process for lesion shape detection. This choice helps speed up training, especially when working with high-dimensional ResNet feature vectors, as linear SVMs are computationally efficient and can handle large feature spaces effectively.

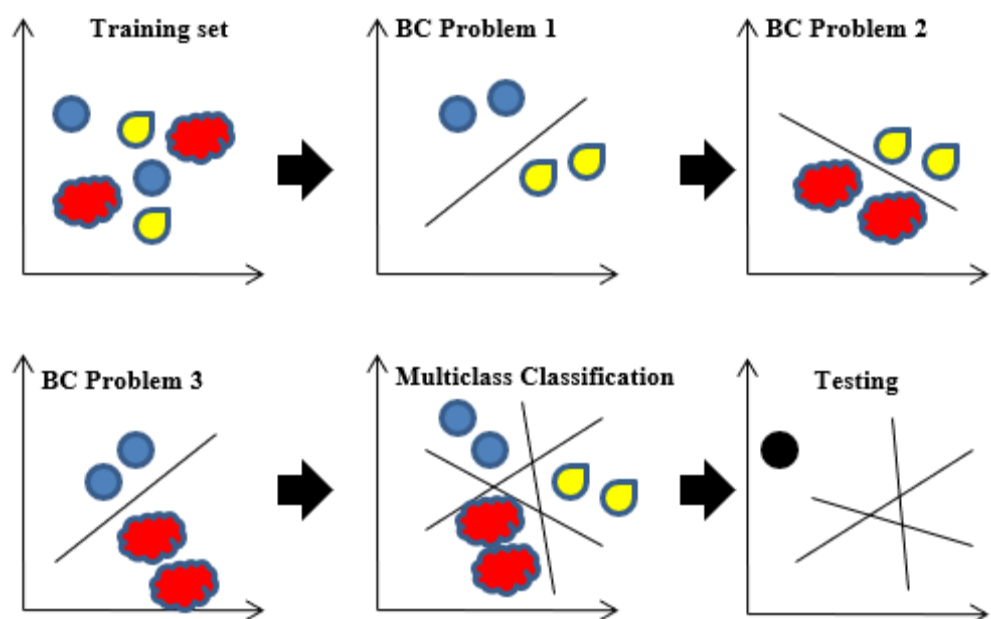

**Fig. 5.** Shape classification using multiclass SVMs with one-versus-one strategy.

### 4. Results and Discussion

### 4.1 Evaluation of Classifier

The above-mentioned procedure is repeated to extract features from the test sets (5% images of both datasets) using ResNet18 (R18), ResNet50 (R50), and ResNet101 (R101). The test-sets are completely unseen data and do not involve in training. The test-features are passed to the multiclass SVMs to measure the accuracy of the trained classifier (SVM). Evaluation process is divided into four steps such as- i) training on ISIC2016 and testing on ISIC2016, ii) training on APH2 and testing on ISIC2016, iii) training on APH2 and testing on APH2, and iv) training on ISIC2016 and testing on APH2, to avoid overfitting and make sure proposed CNN model is trained well. These four steps' outputs are presented respectively in **Fig. 6**, **Fig. 7**, **Fig. 8**, and **Fig. 9** as Confusion Metrics (CMs). Evaluation metrics, Eq. (1) to Eq. (6) are applied on these CMs and results are reported in **Table 6**, **Table 7**, **Table 8**, and **Table 9** respectively. **(c20)**

|  | Testing Set | | |
|---|---|---|---|
| TARGET / OUTPUT | Asymmetric | Half-Symmetric | Symmetric |
| Asymmetric | 32 / 50.79% | 0 / 0.00% | 1 / 1.59% |
| Half-Symmetric | 1 / 1.59% | 16 / 25.40% | 0 / 0.00% |
| Symmetric | 3 / 4.76% | 0 / 0.00% | 10 / 15.87% |

(a) R18+SVM

|  | Testing Set | | |
|---|---|---|---|
| TARGET / OUTPUT | Asymmetric | Half-Symmetric | Symmetric |
| Asymmetric | 32 / 50.79% | 1 / 1.59% | 0 / 0.00% |
| Half-Symmetric | 2 / 3.17% | 15 / 23.81% | 0 / 0.00% |
| Symmetric | 1 / 1.59% | 0 / 0.00% | 12 / 19.05% |

(b) R50+SVM

|  | Testing Set | | |
|---|---|---|---|
| TARGET / OUTPUT | Asymmetric | Half-Symmetric | Symmetric |
| Asymmetric | 31 / 49.21% | 1 / 1.59% | 1 / 1.59% |
| Half-Symmetric | 1 / 1.59% | 16 / 25.40% | 0 / 0.00% |
| Symmetric | 1 / 1.59% | 0 / 0.00% | 12 / 19.05% |

(c) R101+SVM

**Fig. 6.** (a), (b), and (c) are the CMs of ResNets+SVM, when training on ISIC2016 and testing on ISIC2016.

|  | Testing Set | | |
|---|---|---|---|
| TARGET / OUTPUT | Asymmetric | Half-Symmetric | Symmetric |
| Asymmetric | 31 / 49.21% | 0 / 0.00% | 2 / 3.17% |
| Half-Symmetric | 3 / 4.76% | 14 / 22.22% | 0 / 0.00% |
| Symmetric | 2 / 3.17% | 1 / 1.59% | 10 / 15.87% |

(a) R18+SVM

|  | Testing Set | | |
|---|---|---|---|
| TARGET / OUTPUT | Asymmetric | Half-Symmetric | Symmetric |
| Asymmetric | 28 / 44.44% | 3 / 4.76% | 2 / 3.17% |
| Half-Symmetric | 0 / 0.00% | 17 / 26.98% | 0 / 0.00% |
| Symmetric | 3 / 4.76% | 0 / 0.00% | 10 / 15.87% |

(b) R50+SVM

|  | Testing Set | | |
|---|---|---|---|
| TARGET / OUTPUT | Asymmetric | Half-Symmetric | Symmetric |
| Asymmetric | 29 / 46.03% | 3 / 4.76% | 1 / 1.59% |
| Half-Symmetric | 1 / 1.59% | 16 / 25.40% | 0 / 0.00% |
| Symmetric | 1 / 1.59% | 0 / 0.00% | 12 / 19.05% |

(c) R101+SVM

**Fig. 7.** (a), (b), and (c) are the CMs of ResNets+SVM, when Training on APH2 and testing on ISIC2016.

|  | Testing Set | | |
|---|---|---|---|
| TARGET / OUTPUT | Asymmetric | Half-Symmetric | Symmetric |
| Asymmetric | 6 / 19.35% | 2 / 6.45% | 0 / 0.00% |
| Half-Symmetric | 0 / 0.00% | 4 / 12.90% | 1 / 3.23% |
| Symmetric | 0 / 0.00% | 0 / 0.00% | 18 / 58.06% |

(a) R18+SVM

|  | Testing Set | | |
|---|---|---|---|
| TARGET / OUTPUT | Asymmetric | Half-Symmetric | Symmetric |
| Asymmetric | 8 / 25.81% | 0 / 0.00% | 0 / 0.00% |
| Half-Symmetric | 0 / 0.00% | 4 / 12.90% | 1 / 3.23% |
| Symmetric | 0 / 0.00% | 1 / 3.23% | 17 / 54.84% |

(b) R50+SVM

|  | Testing Set | | |
|---|---|---|---|
| TARGET / OUTPUT | Asymmetric | Half-Symmetric | Symmetric |
| Asymmetric | 8 / 25.81% | 0 / 0.00% | 0 / 0.00% |
| Half-Symmetric | 0 / 0.00% | 4 / 12.90% | 1 / 3.23% |
| Symmetric | 0 / 0.00% | 0 / 0.00% | 18 / 58.06% |

(c) R101+SVM

**Fig. 8.** (a), (b), and (c) are the CMs of ResNets+SVM, when training on APH2 and testing on APH2.

|  | Testing Set | | |
|---|---|---|---|
| TARGET / OUTPUT | Asymmetric | Half-Symmetric | Symmetric |
| Asymmetric | 8 / 25.81% | 0 / 0.00% | 0 / 0.00% |
| Half-Symmetric | 0 / 0.00% | 5 / 16.13% | 0 / 0.00% |
| Symmetric | 3 / 9.68% | 2 / 6.45% | 13 / 41.94% |

(a) R18+SVM

|  | Testing Set | | |
|---|---|---|---|
| TARGET / OUTPUT | Asymmetric | Half-Symmetric | Symmetric |
| Asymmetric | 8 / 25.81% | 0 / 0.00% | 0 / 0.00% |
| Half-Symmetric | 0 / 0.00% | 4 / 12.90% | 1 / 3.23% |
| Symmetric | 0 / 0.00% | 2 / 6.45% | 16 / 51.61% |

(b) R50+SVM

|  | Testing Set | | |
|---|---|---|---|
| TARGET / OUTPUT | Asymmetric | Half-Symmetric | Symmetric |
| Asymmetric | 8 / 25.81% | 0 / 0.00% | 0 / 0.00% |
| Half-Symmetric | 0 / 0.00% | 3 / 9.68% | 2 / 6.45% |
| Symmetric | 0 / 0.00% | 0 / 0.00% | 18 / 58.06% |

(c) R101+SVM

**Fig. 9.** (a), (b), and (c) are the CMs of ResNets+SVM, when training on ISIC2016 and testing on APH2.

**Table 6**
Evaluation results from **Fig. 6** (both training and testing on ISIC2016 dataset)

| Model | Precision | | | Recall | | | F1-Score | | | M-F1 | W-F1 | K-Score |
| --- | --- | --- | --- | --- | --- | --- | --- | --- | --- | --- | --- | --- |
| | Asymmetric | Half-Symmetric | Symmetric | Asymmetric | Half-Symmetric | Symmetric | Asymmetric | Half-Symmetric | Symmetric | | | |
| R18+SVM | **0.97** | **0.94** | 0.77 | 0.89 | **1.00** | 0.91 | 0.93 | **0.97** | 0.83 | 0.91 | 0.92 | 0.87 |
| R50+SVM | **0.97** | 0.88 | **0.92** | 0.91 | 0.94 | **1.00** | **0.94** | 0.91 | **0.96** | **0.94** | **0.94** | 0.89 |
| R101+SVM | 0.94 | **0.94** | **0.92** | **0.94** | 0.94 | 0.92 | **0.94** | 0.94 | 0.92 | 0.93 | **0.94** | **0.90** |

**Table 7**
Evaluation results from **Fig. 7** (training on APH2 and testing on ISIC2016 datasets)

| Model | Precision | | | Recall | | | F1-Score | | | M-F1 | W-F1 | K-Score |
| --- | --- | --- | --- | --- | --- | --- | --- | --- | --- | --- | --- | --- |
| | Asymmetric | Half-Symmetric | Symmetric | Asymmetric | Half-Symmetric | Symmetric | Asymmetric | Half-Symmetric | Symmetric | | | |
| R18+SVM | **0.94** | 0.82 | 0.77 | 0.86 | **0.93** | 0.83 | 0.90 | 0.88 | 0.80 | 0.86 | 0.87 | 0.79 |
| R50+SVM | 0.85 | **1.00** | 0.77 | 0.90 | 0.85 | 0.83 | 0.88 | **0.92** | 0.80 | 0.86 | 0.87 | 0.79 |
| R101+SVM | 0.88 | 0.94 | **0.92** | **0.94** | 0.84 | **0.92** | **0.91** | 0.89 | **0.92** | **0.91** | **0.90** | **0.85** |

**Table 8**
Evaluation results from **Fig. 8** (both training and testing on APH2 dataset).

| Model | Precision | | | Recall | | | F1-Score | | | M-F1 | W-F1 | K-Score |
| --- | --- | --- | --- | --- | --- | --- | --- | --- | --- | --- | --- | --- |
| | Asymmetric | Half-Symmetric | Symmetric | Asymmetric | Half-Symmetric | Symmetric | Asymmetric | Half-Symmetric | Symmetric | | | |
| R18+SVM | 0.75 | **0.80** | **1.00** | **1.00** | 0.67 | 0.95 | 0.86 | 0.73 | **0.97** | 0.85 | 0.90 | 0.83 |
| R50+SVM | **1.00** | **0.80** | 0.94 | **1.00** | 0.80 | 0.94 | **1.00** | 0.80 | 0.94 | 0.91 | 0.94 | 0.89 |
| R101+SVM | **1.00** | **0.80** | **1.00** | **1.00** | **1.00** | 0.95 | **1.00** | **0.89** | **0.97** | **0.95** | **0.97** | **0.94** |

**Table 9**
Evaluation results from **Fig. 9** (training on ISIC2016 and testing on APH2 datasets).

| Model | Precision | | | Recall | | | F1-Score | | | M-F1 | W-F1 | K-Score |
| --- | --- | --- | --- | --- | --- | --- | --- | --- | --- | --- | --- | --- |
| | Asymmetric | Half-Symmetric | Symmetric | Asymmetric | Half-Symmetric | Symmetric | Asymmetric | Half-Symmetric | Symmetric | | | |
| R18+SVM | **1.00** | **1.00** | 0.72 | 0.73 | 0.71 | **1.00** | 0.84 | **0.83** | 0.84 | 0.84 | 0.84 | 0.74 |
| R50+SVM | **1.00** | 0.80 | 0.89 | **1.00** | 0.67 | 0.94 | **1.00** | 0.73 | 0.91 | 0.88 | 0.90 | 0.83 |
| R101+SVM | **1.00** | 0.60 | **1.00** | **1.00** | **1.00** | 0.90 | **1.00** | 0.75 | **0.95** | **0.90** | **0.94** | **0.88** |

All the tables compare the performance of R18+SVM, R50+SVM, and R101+SVM models across multiple metrics, including Precision, Recall, F1-Score, M-F1, W-F1, and K-Score, in classifying images into Asymmetric, Half-Symmetric, and Symmetric categories. Across the board, models augmented with SVM exhibit consistently high precision and recall for all classes, indicating their proficiency in correctly identifying these categories.

**Table 6** presents the evaluation results from **Fig. 6**, where both training and testing are conducted on the ISIC2016 dataset. R50+SVM shows high precision and recall for Asymmetric and Symmetric classes, while R18+SVM excels in the Half-Symmetric class. Overall, R101+SVM exhibits superior performance, as indicated by its higher W-F1 (0.94) and K-Score (0.90) compared to the other models, suggesting robustness and accuracy in classifying lesion images in the ISIC2016 dataset.

**Table 7** displays the evaluation outcomes from **Fig. 7**, where models are trained on the APH2 dataset and subsequently tested on the ISIC2016 dataset. R50+SVM demonstrates remarkable precision and recall for Half-Symmetric, indicating its strength in accurately identifying this class. R101+SVM, on the other hand, exhibits the highest F1-Score overall and achieves balanced performance across all classes, as evidenced by its leading M-F1 (0.91) and W-F1 (0.90). R18+SVM maintains competitive performance, albeit slightly trailing R101+SVM and R50+SVM in some metrics. These results underscore R101+SVM as the top-performing model in this scenario, with strong agreement with ground truth labels across all models. Overall, training on APH2 and testing on ISIC2016 yields models with robust classification capabilities, particularly highlighting the efficacy of R101+SVM.

**Table 8** showcases the evaluation outcomes from **Fig. 8**, where models are trained on the APH2 dataset and subsequently tested on the same dataset. R18+SVM and R50+SVM achieve perfect precision, however, slightly lower recall compared to R101+SVM, indicating differences in performance for the Half-Symmetric class. R101+SVM emerges as the top-performing model, demonstrating balanced performance across all metrics, while R50+SVM also performs commendably, especially excelling in precision for all classes. These results highlight the effectiveness of combining ResNet models with SVM augmentation for lesion classification tasks, particularly when training and testing are conducted on the same dataset like APH2.

**Table 9** presents the evaluation results from **Fig. 9**, where models are trained on the ISIC2016 dataset and subsequently tested on the APH2 dataset. All models demonstrate perfect precision for the Asymmetric class, indicating accurate classification. However, there are differences in recall and F1-Score across the Half-Symmetric and Symmetric classes, with R18+SVM achieving the highest F1-Score (0.83) for the Half-Symmetric class. R101+SVM emerges as the top-performing model overall, showing balanced performance across all metrics and strong agreement with the ground truth labels. These results underscore the effectiveness of combining ResNet models with SVM augmentation for lesion classification tasks, particularly when models are trained on one dataset and tested on another, as demonstrated in this scenario.

The analysis across all tables demonstrates the effectiveness of combining ResNet models with SVM augmentation for lesion classification tasks. R101+SVM consistently emerges as the top-performing model, exhibiting balanced performance across various metrics and datasets. These findings suggest that R101+SVM is robust and accurate for classifying skin lesions, making it a suitable model to be deployed for the assessment of asymmetry in skin lesions, which benefits practitioners and researchers in dermatology.

### 4.2 Comparative Study

At present, to the best of our knowledge, there are a few similar approaches found in the literature to compare this shape classification with the proposed method. However, we analyze the existing methods, their experimental datasets (focusing on the total number of images used), the achieved accuracy, and the number of classes. **Table 10** shows the results of different approaches to compare with the proposed one. Several random dermoscopic images from ISIC2018 [23, 24] were used in [11] where asymmetry, color variegation, and diameter features were extracted. Only asymmetry features are given focus in our comparative study. The Kullback-Leibler Divergence technique results were reported as the proposed methods' evaluation metrics in [8]. Then, a feed-forward neural network with Levenberg-Marquardt Backpropagation training algorithm-based experimental results was reported as the regression coefficient and mean squared error in [12]. The dermatological asymmetry measure in a hue based on the threshold binary masks algorithm was presented in [9], and 33 out of 200 images of PH2 were reported as part of a lesion (not a complete lesion image). A CNN-based approach was trained on an augmented PH2 (APH2) dataset and a new dataset SymDerm [14]. Since the proposed CNN (R101+SVM) has slightly better performance than the other two, according to tables 6 to 9, R101+SVM is included in the comparative study.

**Table 10**
A comparative study of existing methods with the proposed method.

| Methods | Number of data | Average Recall (%) | Average Precision (%) | Accuracy (%) | Class |
|---|---|---|---|---|---|
| Decision Tree [11] | 204 [ISIC2018] | - | - | 80.0 | 2 |
| Kullback-Leibler [8] | 200 [PH2] | 81.0 | 79.0 | 80.7 | 3 |
| FFN-LMBP [12] | 80 [PH2] | - | - | 83 | 2 |
| | 80 [Med-Node] | | | 89 | 2 |
| Hue distribution [9] | 167 [PH2] | - | - | 83.2 | 3 |
| CNN [14] | 438[APH2] | 75.1 | 77.0 | 74.8 | 2 |
| | 615[SymDerm] | 59.7 | 53.1 | 42.1 | 3 |
| **Proposed CNN (R101+SVM)** | 600 [APH2] | **90.0** | **91.4** | **90.5** | **3** |
| | 1279 [ISIC2016] From tables 7&9 | **96.7** | **86.7** | **93.5** | **3** |

The proposed method outperforms these five methods in all evaluation metrics. In most of these methods, lesion shape classification was part of a group of feature analysis processes toward recognizing Melanoma. For this reason, asymmetry analysis was not the priority. This could be a potential reason why existing methods from the literature showed lower performance than the proposed method in this research. It is also worth noting that the training environment for these methods was not the same as the proposed method.

To do more comparison under the same environment on APH2 and ISIC2016 datasets, we trained three popular CNNs- AlexNet, GoogLeNet, and SqueezeNet with SVM same as the proposed CNN to classify skin lesion shape. **Table 11** shows the comparison between these three CNNs and the proposed CNN.

**Table 11**
Another comparative study between the proposed CNN and three popular CNNs.

| Methods | Dataset | Average Recall (%) | Average Precision (%) | Accuracy (%) | K-Score |
|---|---|---|---|---|---|
| **Proposed CNN (R101+SVM)** | APH2 | **90.0** | **91.4** | **90.5** | **84.6** |
| | ISIC2016 | **96.7** | **86.7** | **93.5** | **88.1** |
| **AlexNet+SVM** | APH2 | 85.7 | 85.9 | 85.7 | 76.9 |
| | ISIC2016 | 83.2 | 80.6 | 87.1 | 77.4 |
| **GoogLeNet+SVM** | APH2 | 84.3 | 84.9 | 84.1 | 74.5 |
| | ISIC2016 | 89.8 | 78.2 | 87.1 | 76.6 |
| **SqueezeNet+SVM** | APH2 | 81.6 | 82.3 | 82.5 | 71.9 |
| | ISIC2016 | 77.8 | 76.3 | 83.9 | 71.9 |

After the comparison of different evaluation metrics, almost the same experimental results are found in **Table 11**, which indicates the flexibility and reliability of the proposed CNN. In contrast, AlexNet+SVM, GoogLeNet+SVM, and SqueezeNet+SVM exhibit lower performance across all metrics, with decreasing values as compared to the proposed CNN. These results underscore the robustness and efficacy of the proposed CNN model (R101+SVM) for skin lesion classification tasks, highlighting its potential for practical implementation in dermatology applications. Improving accuracy in terms of real-time patient data is a future direction for this research.

## 5. Conclusion

The analysis of lesion shapes holds significant importance in clinical diagnosis, providing vital information for detecting skin diseases. While the value of individual shape analysis may appear limited, it serves as a crucial indicator of underlying skin conditions. In this study, the ISIC2016 dataset was labeled based on clinical experts' opinions. An innovative strategy employing a geometric approach was introduced to identify dermatological symmetry types in the segmented disease area. This method, utilizing binary images, serves as a valuable tool for understanding lesion shapes, particularly for non-experts, and provides essential ground truth data for unlabeled datasets like ISIC2016, offering support for clinical opinions to avoid disagreements among experts. Subsequently, we developed and trained a convolutional neural network (CNN) specifically designed to categorize lesion shapes into three distinct classes. The proposed CNN achieved an outstanding success rate, averaging over 90% compared to conventional automated methods. These findings have substantial implications, significantly enhancing the early-stage diagnosis of skin diseases, including critical conditions like melanoma. This study advances the field of dermatological research, offering a promising avenue for improved diagnosis and understanding of various skin conditions.

In this research, the current approach splits lesion binary images into four parts by 90-degree angles at the centroid and calculates the ratio of white pixels between each part. For more precise analysis, the images should be split into eight parts by 45-degree angles at the centroid, and the ratio of white pixels between each part should be calculated. Another limitation involves the simultaneous consideration of the lesion's border, structure, and color during segmentation. Future studies should examine these three properties separately to gain a deeper understanding of the shapes of asymmetric and symmetric lesions. Evaluating each shape property individually and combining their individual scores for comparison with the proposed algorithm is essential for advancing this research.

Future work should address these identified limitations and further explore the individual and combined analysis of lesion properties (such as pigment network, dots-globules, and lesion colors) to refine and enhance diagnostic accuracy. By focusing on these aspects, future research can continue to improve the methodologies and tools available for clinical diagnosis, ultimately benefiting patient outcomes.

**Conflicts of Interest:** The authors declare that they have no conflicts of interest to report regarding the present study.

**CRediT authorship contribution statement**

**M. A. Rasel:** Conceptualization, Methodology, Software, Validation, Writing – Original Draft, Visualization. **Sameem Abdul Kareem:** Conceptualization, Formal analysis, Investigation, Writing - Review & Editing, Supervision. **Zhenli Kwan:** Writing - Review & Editing, Validation, Formal analysis, Data Curation. **Nik Aimee Azizah Faheem:** Validation, Formal analysis, Data Curation. **Winn Hui Han:** Validation, Formal analysis, Data Curation. **Rebecca Kai Jan Choong:** Validation, Formal analysis, Data Curation. **Shin Shen Yong:** Validation, Formal analysis, Data Curation. **Unaizah Obaidellah:** Conceptualization, Writing- Reviewing and Editing, Supervision, Project administration.

**References**
1. Rigel, D.S., Friedman, R.J., Kopf, A.W., Polsky, D., 2005. ABCDE—An Evolving Concept in the Early Detection of Melanoma. Archives of Dermatology 141. https://doi.org/10.1001/archderm.141.8.1032
2. Zalaudek, I., Argenziano, G., Soyer, H.P., Corona, R., Sera, F., Blum, A., Braun, R.P., Cabo, H., Ferrara, G., Kopf, A.W., Langford, D., Menzies, S.W., Pellacani, G., Peris, K., Seidenari, S., 2005. Three-point checklist of dermoscopy: an open internet study. British Journal of Dermatology 154, 431–437. https://doi.org/10.1111/j.1365-2133.2005.06983.x
3. Henning, J.S., Stein, J.A., Yeung, J., Dusza, S.W., Marghoob, A.A., Rabinovitz, H.S., Polsky, D., Kopf, A.W., 2008. CASH Algorithm for Dermoscopy Revisited. Archives of Dermatology 144. https://doi.org/10.1001/archderm.144.4.554
4. Gutman, David; Codella, Noel C. F.; Celebi, Emre; Helba, Brian; Marchetti, Michael; Mishra, Nabin; Halpern, Allan. "Skin Lesion Analysis toward Melanoma Detection: A Challenge at the International Symposium on Biomedical Imaging (ISBI) 2016, hosted by the International Skin Imaging Collaboration (ISIC)". eprint arXiv:1605.01397. 2016.

**Accepted Manuscript**: This is the peer-reviewed version of the article accepted for publication in *Computers in Biology and Medicine*.
Final published version: https://doi.org/10.1016/j.compbiomed.2024.108851© 2024. This manuscript version is made available under the CC BY-NC-ND license.
5. Lorentzen, H.F., Weismann, K., Larsen, F.G., 2001. Structural asymmetry as a dermatoscopic indicator of malignant melanoma – a latent class analysis of sensitivity and classification errors. Melanoma Research 11, 495–501. https://doi.org/10.1097/00008390-200110000-00009
6. Ng, V.T.Y., Fung, B.Y.M., Lee, T.K., 2005. Determining the asymmetry of skin lesion with fuzzy borders. Computers in Biology and Medicine 35, 103–120. https://doi.org/10.1016/j.compbiomed.2003.11.004
7. 7.Sirakov, N.M., Mete, M., Chakrader, N.S., 2011. Automatic boundary detection and symmetry calculation in dermoscopy images of skin lesions. 2011 18th IEEE International Conference on Image Processing. https://doi.org/10.1109/icip.2011.6115757
8. Chakravorty R, Liang S, Abedini M, Garnavi R. Dermatologist-like feature extraction from skin lesion for improved asymmetry classification in PH2 database. Proceedings of the Annual International Conference of the IEEE Engineering in Medicine and Biology Society, EMBS, vol. 2016-October, Institute of Electrical and Electronics Engineers Inc.; 2016, p. 3855–8. https://doi.org/10.1109/EMBC.2016.7591569.
9. P. Milczarski, Z. Stawska and P. Maślanka, "Skin lesions dermatological shape asymmetry measures," 2017 9th IEEE International Conference on Intelligent Data Acquisition and Advanced Computing Systems: Technology and Applications (IDAACS), 2017, pp. 1056-1061, doi: 10.1109/IDAACS.2017.8095247.
10. Sancen-Plaza, A., Santiago-Montero, R., Sossa, H., Perez-Pinal, F.J., Martinez-Nolasco, J.J., Padilla-Medina, J.A., 2018. Quantitative evaluation of binary digital region asymmetry with application to skin lesion detection. BMC Medical Informatics and Decision Making 18. https://doi.org/10.1186/s12911-018-0641-7
11. Ali AR, Li J, O'Shea SJ. Towards the automatic detection of skin lesion shape asymmetry, color variegation and diameter in dermoscopic images. PLoS One 2020;15. https://doi.org/10.1371/journal.pone.0234352.
12. Damian FA, Moldovanu S, Moraru L. Skin Lesions Asymmetry Estimation Using Artificial Neural Networks. 2021 25th International Conference on System Theory, Control and Computing, ICSTCC 2021 - Proceedings, Institute of Electrical and Electronics Engineers Inc.; 2021, p. 64–7. https://doi.org/10.1109/ICSTCC52150.2021.9607133.
13. Zhang, G., Guo, S., 2021. Asymmetry analysis of melanoma based on ABCD rule. Journal of Physics: Conference Series 1883, 012070. https://doi.org/10.1088/1742-6596/1883/1/012070
14. Talavera-Martínez, L., Bibiloni, P., Giacaman, A., Taberner, R., Hernando, L.J.D.P., González-Hidalgo, M., 2022. A novel approach for skin lesion symmetry classification with a deep learning model. Computers in Biology and Medicine 145, 105450. https://doi.org/10.1016/j.compbiomed.2022.105450
15. Teresa Mendonça, Pedro M. Ferreira, Jorge Marques, Andre R. S. Marcal, Jorge Rozeira. PH² - A dermoscopic image database for research and benchmarking, 35th International Conference of the IEEE Engineering in Medicine and Biology Society, July 3-7, 2013, Osaka, Japan.
16. Deng, Jia, et al. "Imagenet: A large-scale hierarchical image database." Computer Vision and Pattern Recognition, 2009. CVPR 2009. IEEE Conference on. IEEE, 2009.
17. He, K., Zhang, X., Ren, S., Sun, J., 2016. Deep Residual Learning for Image Recognition. 2016 IEEE Conference on Computer Vision and Pattern Recognition (CVPR). https://doi.org/10.1109/cvpr.2016.90
18. Krizhevsky, A., Sutskever, I., Hinton, G.E., 2017. ImageNet classification with deep convolutional neural networks. Communications of the ACM 60, 84–90. https://doi.org/10.1145/3065386
19. Alzubaidi, L., Zhang, J., Humaidi, A.J., Al-Dujaili, A., Duan, Y., Al-Shamma, O., Santamaría, J., Fadhel, M.A., Al-Amidie, M., Farhan, L., 2021. Review of deep learning: concepts, CNN architectures, challenges, applications, future directions. Journal of Big Data 8. https://doi.org/10.1186/s40537-021-00444-8
20. Donahue, J., Jia, Y., Vinyals, O., Hoffman, J., Zhang, N., Tzeng, E. and Darrell, T., 2014, January. Decaf: A deep convolutional activation feature for generic visual recognition. In *International conference on machine learning* (pp. 647-655). PMLR.
21. Arefin, M.R., Asadujjaman, M., 2016. Minimizing Average of Loss Functions Using Gradient Descent and Stochastic Gradient Descent. Dhaka University Journal of Science 64, 141–145. https://doi.org/10.3329/dujs.v64i2.54490
22. Sun, P., Reid, M.D., Zhou, J., 2014. An improved multiclass LogitBoost using adaptive-one-vs-one. Machine Learning 97, 295–326. https://doi.org/10.1007/s10994-014-5434-3
23. Noel Codella, Veronica Rotemberg, Philipp Tschandl, M. Emre Celebi, Stephen Dusza, David Gutman, Brian Helba, Aadi Kalloo, Konstantinos Liopyris, Michael Marchetti, Harald Kittler, Allan Halpern: "Skin Lesion Analysis Toward Melanoma Detection 2018: A Challenge Hosted by the International Skin Imaging Collaboration (ISIC)", 2018; https://arxiv.org/abs/1902.03368
24. Tschandl, P., Rosendahl, C., Kittler, H., 2018. The HAM10000 dataset, a large collection of multi-sourcedermatoscopic images of common pigmented skin lesions. Scientific Data 5. https://doi.org/10.1038/sdata.2018.161